\documentclass{article}

\usepackage{spconf,amsmath,graphicx}
\usepackage{xcolor}
\usepackage{url}

\title{Anomalous Motion Detection on Highway Using Deep Learning}

\name{Harpreet Singh \qquad Emily M. Hand \qquad Kostas Alexis}
\address{University of Nevada, Reno, USA \\
        Department of Computer Science and Engineering}

\begin{document}
\fontsize{10}{10}

\maketitle

\begin{abstract}
Research in visual anomaly detection draws much interest due to its applications in surveillance. Common datasets for evaluation are constructed using a stationary camera overlooking a region of interest. Previous research has shown promising results in detecting spatial as well as temporal anomalies in these settings. The advent of self-driving cars provides an opportunity to apply visual anomaly detection in a more dynamic application yet no dataset exists in this type of environment. This paper presents a new anomaly detection dataset -- the Highway Traffic Anomaly (HTA) dataset -- for the problem of detecting anomalous traffic patterns from dash cam videos of vehicles on highways. We evaluate state-of-the-art deep learning anomaly detection models and propose novel variations to these methods. Our results show that state-of-the-art models built for settings with a stationary camera do not translate well to a more dynamic environment. The proposed variations to these SoTA methods show promising results on the new HTA dataset.
\end{abstract}


\section{Introduction}
\begin{figure}[!t]
    \centering
    \includegraphics[width=2.2in]{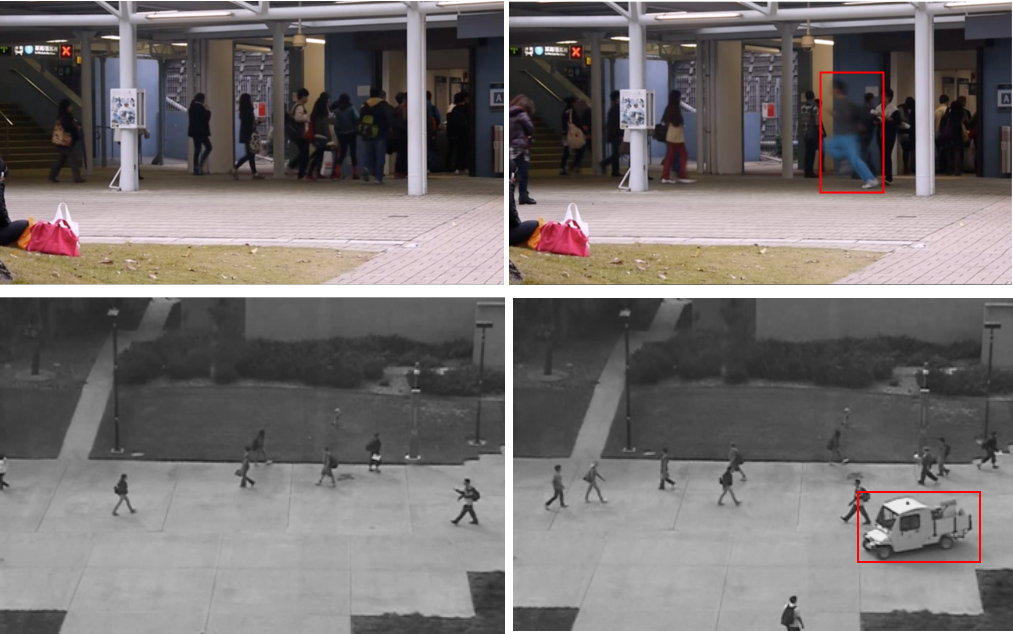}
    \caption{The top row shows examples from the CUHK anomaly dataset, normal image data(left) and abnormal(right). The bottom row is an example from the UCSD anomaly dataset, the anomaly(right) is the golf cart}
    \label{fig:sample-anomaly-data-set}
\end{figure}

Anomaly detection is an unsupervised one-class classification problem with the goal of learning the normal state of data during training and then detecting aberrations without any provided labels. Many anomaly detection applications involve visual data, such as images or videos, and are motivated by interest in surveillance. Datasets commonly evaluated for visual anomaly detection are constructed with a stationary camera observing a region in which the background environment is relatively static while foreground objects such as pedestrians and vehicles are in motion. Anomalies are defined as appearance or motion deviations from normal data in the foreground objects, shown in Figure \ref{fig:sample-anomaly-data-set}. \cite{Patil2016ASO} provides a comprehensive list of video anomaly detection datasets, but they all maintain the same characteristics. As autonomous robots become increasingly common, there is a need to perform anomaly detection while an agent is moving within an environment yet no such dataset exists.

This paper presents a new anomaly detection dataset, the Highway Traffic Anomaly (HTA) dataset, to evaluate anomaly detection methods with a moving agent. Specifically, the dataset consists of dash cam videos captured from vehicles driving on the highway. The goal is to learn normal driving motion of traffic in the camera's field of view and then detect conditions in which other vehicles are moving abnormally. In this dataset, not only is the camera moving but other vehicles and background features are also in motion relative to the camera. The dataset consists of five types of anomalies: speeding vehicle, speeding motorcycle, vehicle accident, close merging vehicle and halted vehicle. Three state-of-the-art deep learning based anomaly detection models are evaluated and two variations, specifically for the problem of detecting anomalous highway traffic motion, are proposed. Code for the HTA dataset and the evaluated models is available at \cite{github-this-project}.

\section{Related Work}
\label{section:related-work}

\subsection{Anomaly Detection Datasets}
Frequently evaluated visual anomaly datasets consist of a static background, moving foreground objects, and a stationary camera. The UCSD Pedestrian dataset \cite{mahadevan2010anomaly} consists of videos of pedestrians on a university campus. Normal data corresponds to pedestrians walking on pathways. Abnormal data consists of small vehicles or cyclists that are different in appearance and motion from usual pedestrian traffic. The CUHK \cite{lu2013abnormal} Avenue dataset also consists of videos of a crowded walkway with anomalies defined as unusual behavior, such as throwing an object. Similarly, the UMN Unusual Crowd Activity dataset \cite{umn-crowd-data} consists of videos of unusual behavior in crowds. While not exhaustive, this list summarizes the typical visual anomaly datasets previously studied. To the best of our knowledge, there is no open-source autonomous driving dataset specifically for the task of anomaly detection.

\subsection{Anomaly Detection Methods}
Initial work in anomaly detection relied on hand-engineered features (e.g., HOG) to create meaningful representations of normal image data. The features were then used to fit a statistical model such as an SVDD \cite{tax2004support}. Deep learning models have recently shown promising results in anomaly detection. The deep learning models evaluated in this study can be grouped using the taxonomy proposed in \cite{kiran2018-taxonomy}: generative and predictive learning models.

Generative models attempt to estimate the probability distribution of the training data (i.e. the normal motion data). After training, test videos are evaluated by some form of a reconstruction error. \cite{dimokranitou2017adversarial-anomaly} proposes a method to use adversarial autoencoders for anomaly detection by relying on the generator's ability to model only the normal data distribution. More recently, a conditional generative adversarial network (CGAN) is used by \cite{ravanbakhsh2019-cgan-anomaly}. The authors train two independent CGANs to learn an image-to-flow and flow-to-image transformation. We evaluate this model on the HTA dataset, providing more details in section \ref{section:method}. Predicting Optical flow with CNNs for visual understanding has been shown to be effective, such as in \cite{martin2019optimal} for action classification.

Predictive networks seek to model sequential data. This approach is effective in learning temporal patterns in videos; given a sequence of $N$ video frames, predict the $N+1$ frame. Anomaly detection is performed by first training only with normal data to accurately predict images from normal motion sequences. After training, predicting a future frame of an abnormal sequence will result in a larger error. Various architectures have been proposed \cite{medel2016-lstm-anomaly, chong2017-enc-3dlstm-dec, luo2017remembering-history}, they all share the same anomaly detection mechanism.

\section{Method}
\label{section:method}

\subsection{Highway Traffic Anomaly Dataset}
The HTA dataset was curated from the Berkeley DeepDrive dataset \cite{yu2018-berk-deepdrive} that consists of 100k high resolution($1280 \times 720$, $30 FPS$) dash cam videos collected from cars in New York and the Bay Area. We sifted through the entire dataset, selecting only highway driving videos. From that subset, videos in visually degraded conditions were removed. In summary, the HTA dataset consists of highway videos: during clear lighting conditions; clear, partly cloudy, or overcast weather conditions; minimally occluded from large vehicles; and contain at least some traffic in motion. Due to the imperfect nature of the data collection process, the HTA dataset is not without noise. For instance, bumps and cracks on the highway cause transient shaking in the videos. These characteristics make the HTA dataset more challenging and realistic.

\begin{figure}[!t]
	\centering
	\includegraphics[width=2.0in]{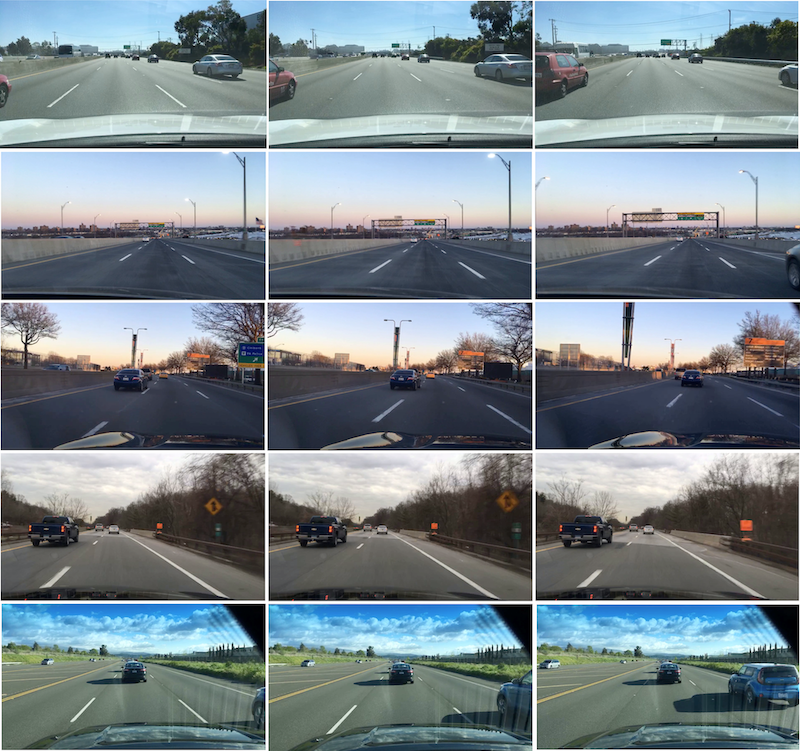}
	\caption{Examples of normal images from the training set.}
	\label{fig:norm-images-fig}
\end{figure}

Normal driving conditions in this dataset are defined by the motion of vehicles that does not perturb the motion of the dash cam vehicle or the motion of other vehicles stays relatively self-similar. The training set consists of $286$ videos of normal traffic conditions, a total of $322,202$ video frames and an average duration of $40$ seconds. Figure \ref{fig:norm-images-fig} shows a sample of video frames from the training set.

\begin{figure}[!t]
	\centering
	\includegraphics[width=2.2in]{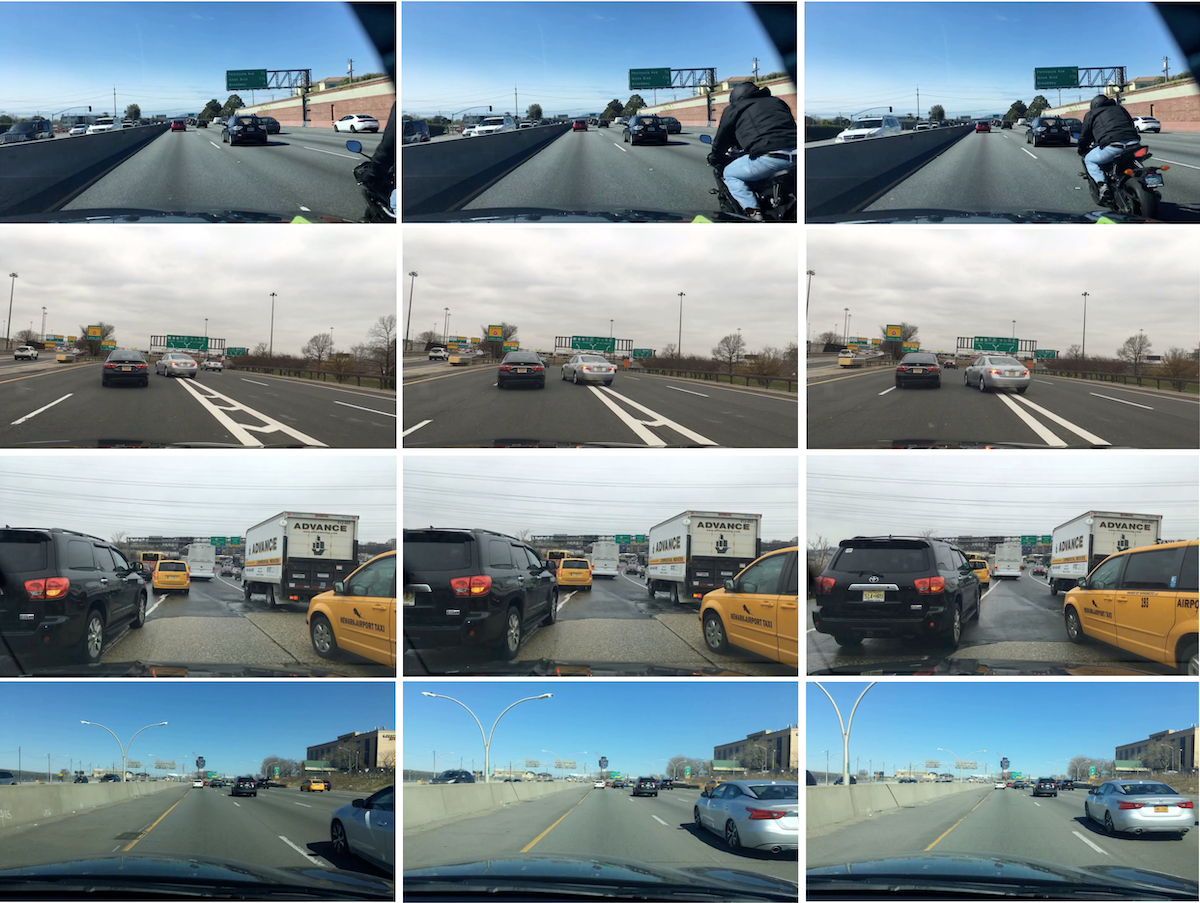}
	\caption{Examples of abnormal images from the training set. Top to bottom row: speeding motorcycle, halted vehicle, close merge, speeding vehicle. Vehicle accident not shown but YouTube links will be provided.}
	\label{fig:abnorm-images-fig}
\end{figure}

The test set contains a total of $103$ videos, $78$ normal traffic videos not in the training set and $25$ abnormal traffic videos. Abnormal traffic videos consist of five types of anomalous motion: speeding vehicle (4 videos), close merge (13 videos), halted vehicle (1 video), vehicle accident (5 videos), and speeding motorcycle (2 videos). Each case represents a situation in which a human driver will practice caution. Vehicle accident anomalies were downloaded from YouTube. Example image sequences of each abnormality are shown in Figure \ref{fig:abnorm-images-fig}. Abnormal motion is manually annotated at the frame level since only short sequences contain abnormal motion. A frame was labelled anomalous if the motion from the previous frame to the current was part of an anomalous motion. There are a total of $1531$ frames labeled as anomalous, making up 6\% of the abnormal test set.

\subsection{Generative Models}
\begin{figure}[!t]
	\centering
	\includegraphics[width=2.1in]{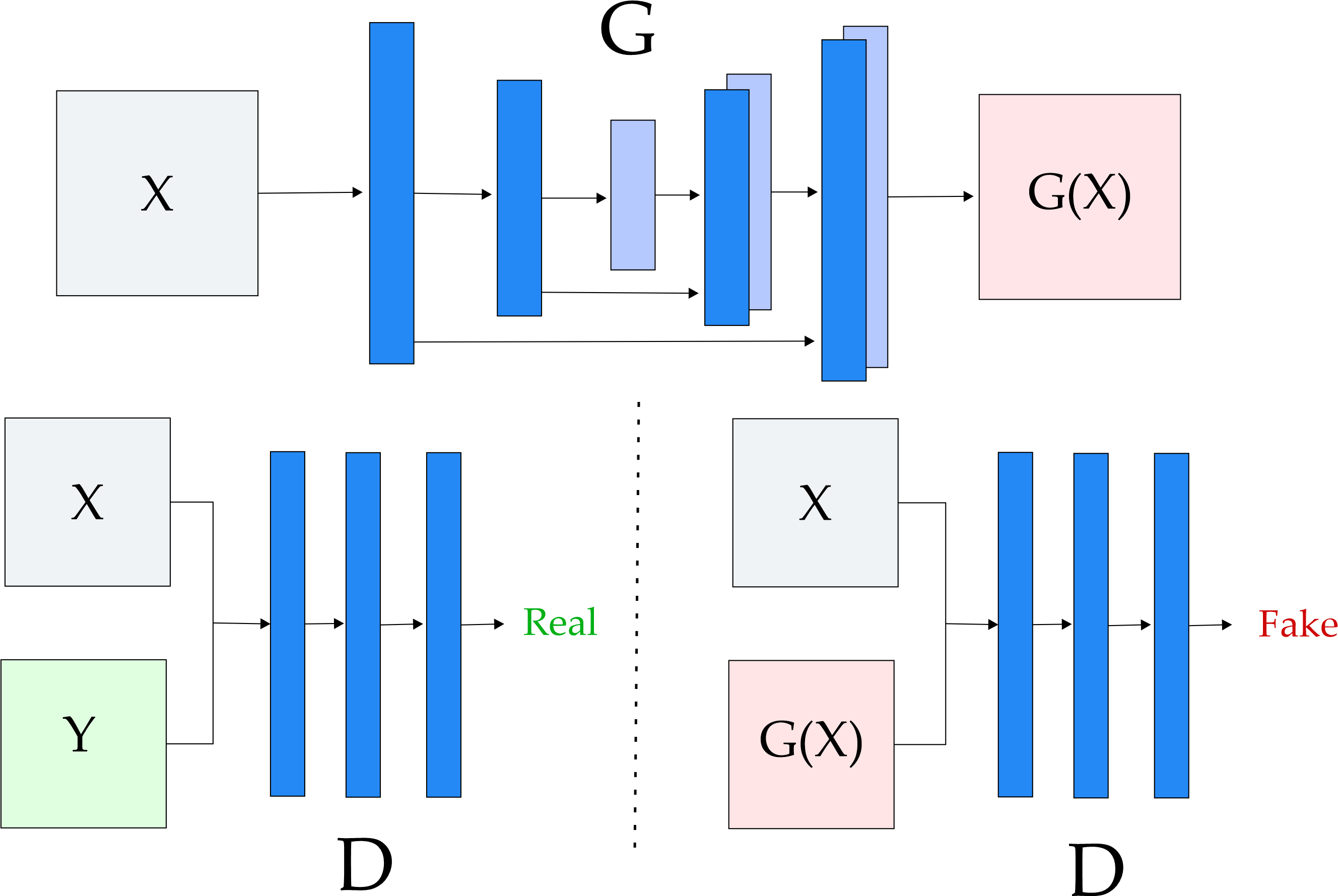}
	\caption{The CGAN's generator learns an image-to-optical flow transformation. The discriminator processes the input and either the ground truth or generator's output concatenated and classify it as real or fake.}
	\label{fig:cgan-model}
\end{figure}

Since an anomaly is defined as irregular motion, generative models can learn to predict dense optical flow to model normal motion. For training, ground truth optical flow is computed using OpenCV's \cite{opencv_library} dense optical flow implementation. The first generative model evaluated is the conditional GAN (CGAN) proposed in \cite{ravanbakhsh2019-cgan-anomaly}, Figure \ref{fig:cgan-model}. The CGAN is trained to predict the optical flow between a pair of sequential frames. The generator's input are two RGB images, concatenated depth-wise and it predicts the corresponding optical flow. The discriminator classifies patches in the input as real or fake, producing a 2D-one channel output in the range $[0, 1]$, 1 denoting real and 0 denoting fake.

\cite{ravanbakhsh2019-cgan-anomaly} proposes using the discriminator's output to detect anomalies since it learns to distinguish normal motion. The discriminator's output can be interpreted as a heat map for patches containing abnormal motion. A pair of images will be classified as anomalous if there exists an element in the discriminator's output below a threshold. The second approach to detect abnormal motion uses the pixel-level difference between generator's predicted optical flow and the ground truth optical flow. The difference is then averaged using a sliding window. A frame is labeled as anomalous if there exists an error above a threshold in either the $x$ or $y$ component. 

Due to the high capacity of GANs, they may estimate anomalous data as well as normal data \cite{kiran2018overview}. In order to evaluate the effectiveness of the CGAN, we also evaluate the state-of-the-art deep learning dense optical flow prediction model: FlowNet \cite{fischer2015-flownet}. Unlike the CGAN, FlowNet is not trained adversarially.

\subsection{Predictive Model}
\begin{figure}[!t]
	\centering
	\includegraphics[width=2.5in]{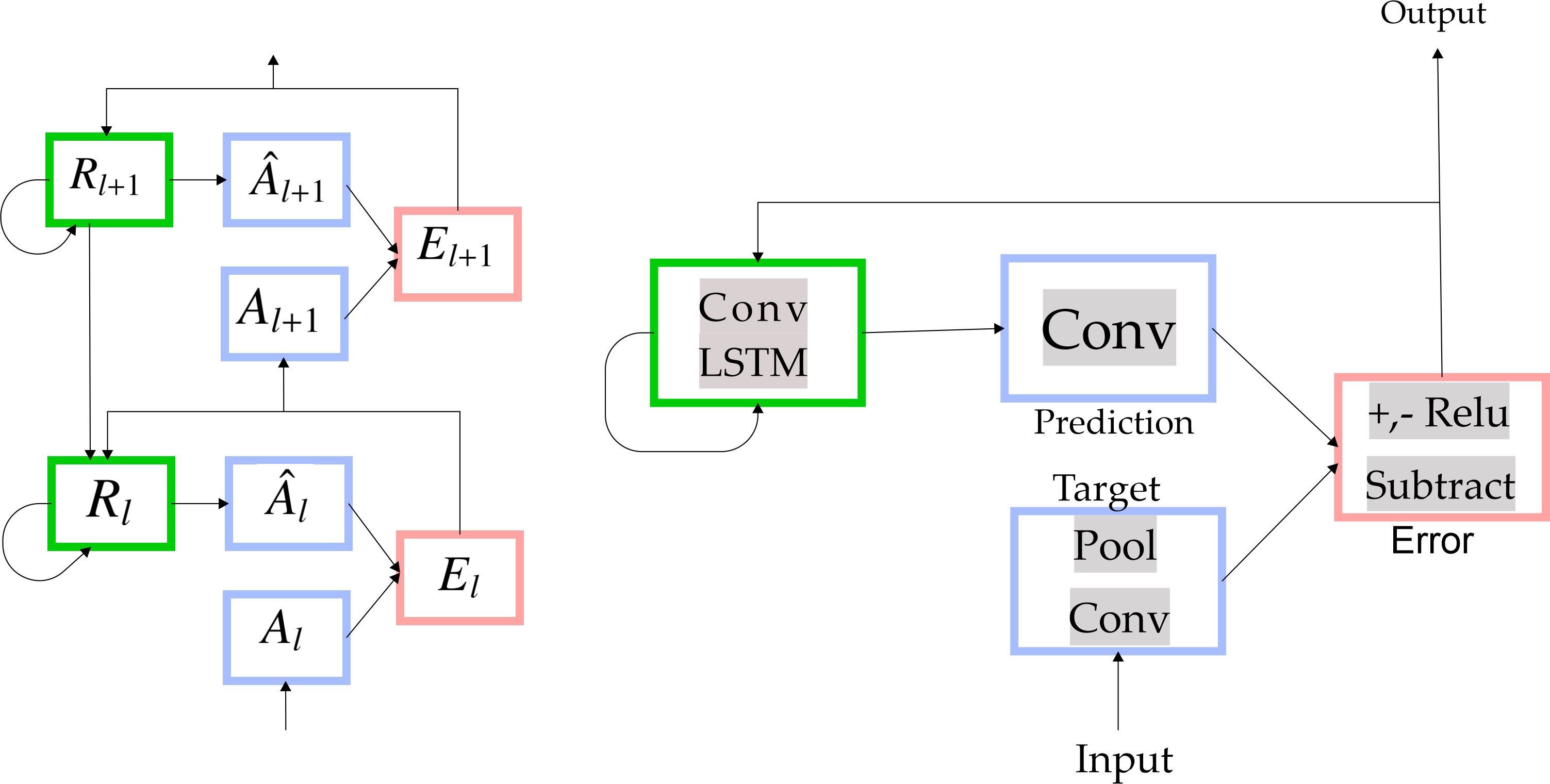}
	\caption{Predictive Coding Network architecture is composed of modules containing four components: recurrent layer, prediction layer, input layer, and error layer.}
	\label{fig:pred-net}
\end{figure}

Predictive models can learn sequential data such as videos. By training on sequences of $N$ normal traffic frames, a look back of $N$, a predictive model will then predict the $N+1$ frame. The Predictive Coding Network (PredNet) proposed in \cite{lotter2016-prednet} showed promising results in predicting future frames in the KITTI dataset. PredNet, shown in Figure \ref{fig:pred-net}, is constructed with a series of modules that perform local predictions and then propagates only the errors between the predicted and actual to the subsequent layers. The evaluated model was constructed using the recommendations in \cite{lotter2016-prednet}. Anomalies are detected using the pixel-level reconstruction error between the predicted $(N+1)^{th}$ frame and the ground truth frame, employing the averaging sliding window approach to compute pixel-level difference.

\subsection{Proposed Variations}
The first modification we propose is related to the detection mechanism that relies on the reconstruction error. The pixel-level reconstruction error averaged with a sliding window can be improved by using a variable window size. Since vehicles in the center of the image tend to be further from the camera, anomalous motion near the center of the image will displace a small number of pixels. In order to account for depth, we propose to use a smaller 3x3 averaging window size near the center and a larger 9x9 averaging window size around the edges.

Another characteristic, specific to highway anomaly detection, is the motion of the vehicles is relative due to egomotion; vehicles moving at speeds similar to the camera will seem to displace very few pixels, if any. When using the PredNet model to predict future frames, rather than just one frame into the future, the model can predict the $6^{th}$ future frame instead. \cite{lotter2016-prednet} tests the PredNet's capabilities of predicting frames further into the future by using PredNet's first predicted frame as input for the next prediction. Five frames further into the future are shown to maintain reasonable accuracy. We evaluated this method using N=1,2,3,... and found that N=6 gave the best results.

\section{Experiments}
\label{section:experiments}
A total of four models are evaluated with the HTA dataset: CGAN, FlowNet, PredNet $N+1$, and PredNet $N+6$. The performance of each model is measured using an AUC score. The thresholds in each experiment range from $0.002$ to $1.0$ in increments of $0.002$. Five AUC scores are reported for each model, one for each type of anomalous motion, Table \ref{tab:auc-scores}.

\begin{table}[!t]
    \fontsize{7}{9}\selectfont
    \centering
    \caption{AUC scores computed from the all the evaluated models. PredNet(N+6) performs the best across the board.}
    \label{tab:auc-scores}
    \begin{tabular}{|l|l|l|l|l|}
        \hline
                            & CGAN  & FlowNet & PredNet(N+1) & PredNet(N+6) \\ \hline
        Speeding Vehicle    & 0.608 & \textbf{0.623}   & 0.497        & 0.614        \\ \hline
        Accident            & 0.607 & \textbf{0.657}   & 0.559        & 0.601        \\ \hline
        Speeding motorcycle & 0.580 & 0.593   & 0.581        & \textbf{0.828}        \\ \hline
        Close Merge         & 0.422 & 0.531   & 0.619        & \textbf{0.643}        \\ \hline
        Halted Vehicle      & 0.337 & 0.216   & \textbf{0.554}        & 0.236        \\ \hline
        \end{tabular}
\end{table}

\subsection{Conditional GAN}
The CGAN is trained on pairs of RGB images. Input images are cropped from the bottom to remove the visible hood of the vehicle as well as from the top to reduce the amount of sky/background in each frame. The final input size is $128\times512\times6$. The model is trained for 40 epochs.

\begin{figure}[!t]
	\centering
	\includegraphics[width=3.0in]{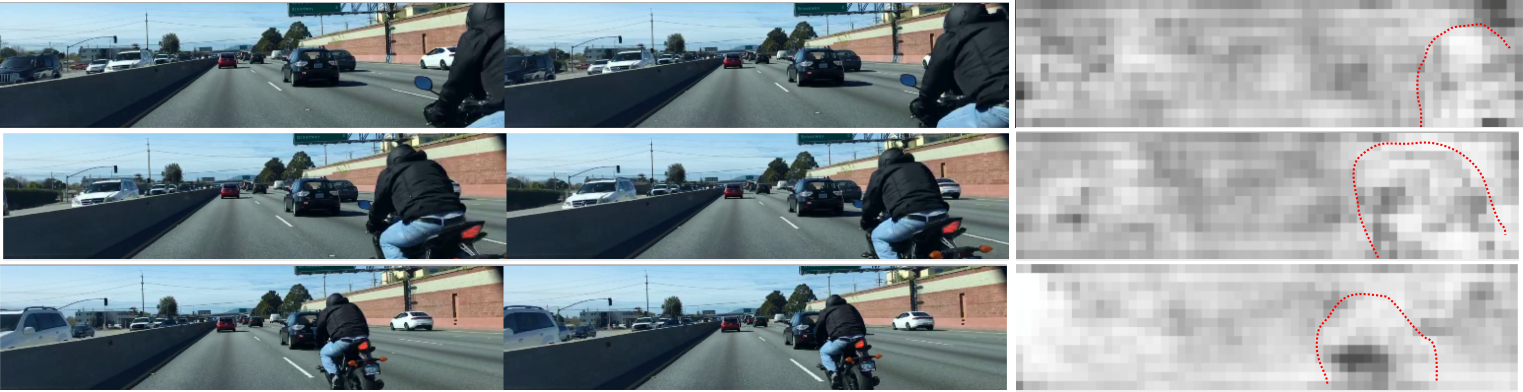}
	\caption{The CGAN discriminator’s output(right) is unable to identify the anomalous motion of the motorcycle.}
	\label{fig:cgan-heat-map}
\end{figure}

The first approach to detect anomalies is by using the discriminator's 2D - one channel output as a heat map for patches containing an anomaly. \cite{ravanbakhsh2019-cgan-anomaly} shows promising results using this approach but the discriminator's output trained on the HTA dataset does not produce the same results, Figure \ref{fig:cgan-heat-map}. The salient characteristic that may cause this discrepancy in the results is that, unlike the UCSD Pedestrian dataset, the HTA dataset does not maintain a static background.

The second approach to detect anomalies with the CGAN is to use the reconstruction error from the generator's output. The AUC scores of the CGAN in Table \ref{tab:auc-scores} show that all anomaly types are near $0.5$, meaning that the reconstruction error from generator's output has minimal discriminative capabilities in classifying abnormal motion.

\subsection{FlowNet}
FlowNet is also trained on pairs of RGB images from the training dataset. The input RGB images for FlowNet are cropped from the bottom, making the input size $2\times256\times512\times3$. The model is trained for 100 epochs. Anomalies are detected in the test set using the same approach as CGAN, the reconstruction error between the predicted and ground truth optical flow. The AUC scores for the FlowNet are only slightly better than the CGAN results, Table \ref{tab:auc-scores}.

Both generative models estimate optical flow of abnormal motion just as well as normal motion; AUC scores of both models are near $0.5$, indicating that they are unable to discern abnormal motion from normal motion. It seems that both models learn to estimate optical flow better than learning distinguishing patterns in normal motion.

\subsection{Predictive Coding Network}
The PredNet model is trained on a look back of 10 consecutive RGB images as suggested by the original work. As with the CGAN, each image is cropped from the bottom as well as the top with the final input size $64\times256\times3\times10$. The model was trained for 100 epochs. Anomalies are detected using the reconstruction error between the model's predicted future frame and the ground truth frame. Using this mechanism, AUC scores for PredNet $N+1$ are provided in Table \ref{tab:auc-scores}. The AUC scores of PredNet $N+1$ show that close merge and speeding motorcycle anomalies perform relatively better than others and achieves the highest AUC score for the halted vehicle anomaly albeit still near 0.5.

\begin{figure}[!t]
	\centering
	\includegraphics[width=3.0in]{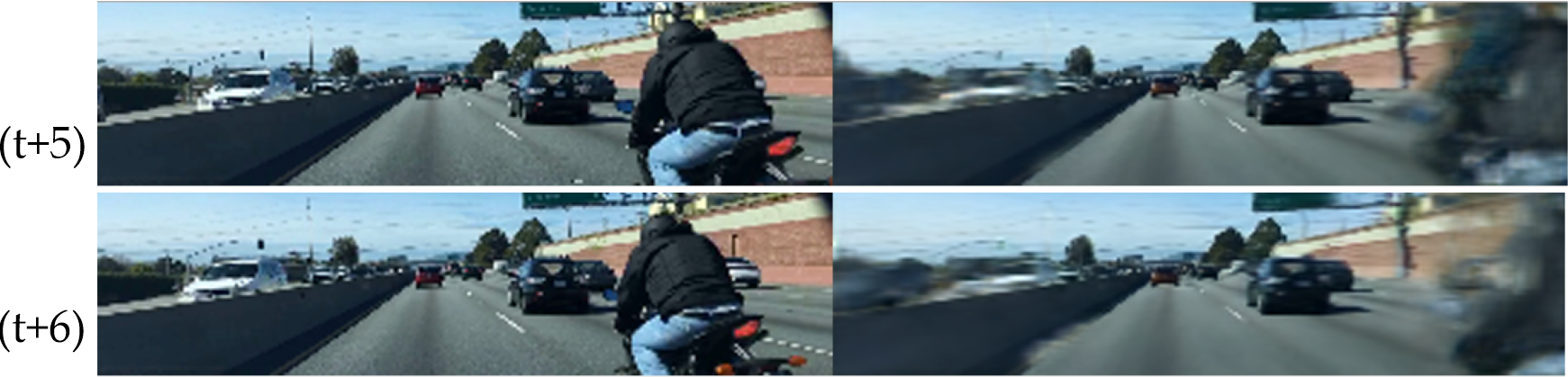}
	\caption{Results from PredNet of the extrapolated frames of abnormal motion, the speeding motorcycle. Vehicles in normal motion are less blurry than the speeding motorcycle.}
	\label{fig:prednet-6-out}
\end{figure}

By setting a look back of two, we extrapolate four frames into the future, testing the sixth frame for anomalous motion, PredNet $N+6$. Figure \ref{fig:prednet-6-out} shows a sample output and ground truth frame. Anomalies are detected in the same manner as with PredNet $N+1$. AUC scores for PredNet $N+6$ are shown in Table \ref{tab:auc-scores} and show significant improvement in detecting the speeding motorcycle anomaly. PredNet $N+6$ achieves the highest or close to the highest AUC score for each anomaly type except for halted vehicle. 

\section{Conclusion}
\label{section:conclusion}
This paper presents a new anomaly detection dataset, the Highway Traffic Anomaly (HTA) dataset. It differs from existing anomaly detection datasets in many ways that make it more challenging. To the best of our knowledge, this is the first anomaly detection dataset for autonomous driving. Four state-of-the-art deep learning models were evaluated with a proposed heuristic to improve the reconstruction error for anomaly detection tailored for the HTA dataset. The results indicate that state-of-the-art models do not perform well on the HTA dataset. Our proposed variation of the PredNet model to predict the sixth future frame shows promising results on the speeding motorcycle anomaly and relatively better in all anomalies.

\newpage

\bibliographystyle{IEEEbib}
\bibliography{bibliography}

\end{document}